\documentclass[journal,onecolumn, draftclsnofoot]{IEEEtran}
\usepackage{graphicx}%
\usepackage{epsfig}
\usepackage{graphicx}%
\usepackage{multirow}%
\usepackage{amsmath,amssymb,amsfonts}%
\usepackage{amsthm}%
\usepackage{mathrsfs}%
\usepackage{xcolor}%
\usepackage{textcomp}%
\usepackage{manyfoot}%
\usepackage{booktabs}%
\usepackage{algorithm}%
\usepackage{algorithmicx}%
\usepackage{algpseudocode}%
\usepackage{listings}%
\usepackage[flushleft]{threeparttable}

\usepackage{setspace}
\begin{document}
%
\title{Digital Modeling on Large Kernel Metamaterial Neural Network}
%
%
%

\author{Quan Liu, Hanyu Zheng, Brandon T. Swartz, Ho hin Lee, Zuhayr Asad,
Ivan Kravchenko \\Jason G. Valentine, Yuankai Huo,
\thanks{Yuankai Huo is the corresponding author, Vanderbilt University, Nashville, TN 37212, USA. E-mail: (yuankai.huo@vanderbilt.edu)}
\thanks{Quan Liu, Hanyu Zheng, Brandon T. Swartz, Ho hin Lee, Zuhayr Asad,
Ivan Kravchenko and Jason G. Valentine are with the Vanderbilt University, Nashville, TN 37212, USA. E-mail: (quan.liu@vanderbilt.edu; hanyu.zheng@vanderbilt.edu; brandon.t.swartz@vanderbilt.edu; ho.hin.lee@vanderbilt.edu; zuhayr.asad@vanderbilt.edu; jason.g.valentine@vanderbilt.edu).}
\thanks{Ivan Kravchenko is with Oak Ridge National Laboratory, Oak Ridge, TN 37830, USA. E-mail: (kravchenkoii@ornl.gov)}

}

\maketitle

\begin{abstract}
Deep neural networks (DNNs) utilized recently are physically deployed with computational units (e.g., CPUs and GPUs). Such a design might lead to a heavy computational burden, significant latency, and intensive power consumption, which are critical limitations in applications such as the Internet of Things (IoT), edge computing, and the usage of drones. Recent advances in optical computational units (e.g., metamaterial) have shed light on energy-free and light-speed neural networks. However, the digital design of the metamaterial neural network (MNN) is fundamentally limited by its physical limitations, such as precision, noise, and bandwidth during fabrication. Moreover, the unique advantages of MNN's (e.g., light-speed computation) are not fully explored via standard 3$\times$3 convolution kernels. In this paper, we propose a novel large kernel metamaterial neural network (LMNN) that maximizes the digital capacity of the state-of-the-art (SOTA) MNN with model re-parametrization and network compression, while also considering the optical limitation explicitly. The new digital learning scheme can maximize the learning capacity of MNN while modeling the physical restrictions of meta-optic. With the proposed LMNN, the computation cost of the convolutional front-end can be offloaded into fabricated optical hardware. The experimental results on two publicly available datasets demonstrate that the optimized hybrid design improved classification accuracy while reducing computational latency. The development of the proposed LMNN is a promising step towards the ultimate goal of energy-free and light-speed AI.
\end{abstract}

\begin{IEEEkeywords}
large convolution kernel, model compression, model re-parameterization, meta-material fabrication adaptation.
\end{IEEEkeywords}

%

\section{Introduction}

Digital neural networks (DNN) are essential in modern computer vision tasks. The convolutional neural network (CNN) is arguably the most widely used AI approach for image classification~\cite{lecun1989backpropagation, krizhevsky2017imagenet, li2014medical}, segmentation~\cite{jha2020doubleu, ronneberger2015u}, and detection~\cite{chauhan2018convolutional, redmon2016you}. Even for more recent Vision Transformer-based models, convolution is still  essential components for extracting local image features~\cite{liu2021swin, wang2021pyramid,liu2022convnet, ding2022scaling, liu2022more}. Current CNNs are typically deployed with computational units (e.g., CPUs and GPUs). Such a design might lead to a heavy computational burden, significant latency, and intensive power consumption, which are critical limitations in applications such as the Internet of Things (IoT), edge computing, and the usage of drones. Therefore, the AI community has started to seek DNN models with less energy consumption and lower latency. However, we might never approach energy-free and light-speed DNN following the current trends in research.

Fortunately, the recent advances in optical computational units (e.g., metamaterial) have shed light on energy-free and light-speed neural networks (Fig.~\ref{fig:fig1}). At its current stage, the SOTA metamaterial neural network (MNN) is implemented as a hybrid system, where the optical processors are used as a light-speed and energy-free front-end convolutional operator with a digital feature aggregator. Such design reduces the computational latency since the convolution operations are implemented by optical units, which off-loads more than 90 percent of the floating-point operations (FLOPs) in conventional CNN backbones like VGG~\cite{simonyan2014very} and ResNet~\cite{he2016deep}. However, the digital design of the MNN is fundamentally limited by its physical structures, namely (1) \textbf{the optic system can only take positive value}; (2) \textbf{non-linear computations are challenging for free-space optic devices at low light intensity}; (3) \textbf{the implementation of the optical convolution is restricted by limited kernel size, channel number, precision, noise, and bandwidth}. Furthermore, limitations also exist in the current optic fabrication process: 1) \textbf{only the first layer of a neural network can be fabricated}, and 2) \textbf{limited layer capacity and weight precision}. Therefore, the unique advantages of the MNNs (e.g., light-speed computation) are not fully explored via standard 3$\times$3 convolution kernels. 

In this paper, we propose a novel large kernel metamaterial neural network (LMNN) that maximizes the digital capacity of the state-of-the-art (SOTA) MNN with model re-parametrization and network compression, while also considering the optical limitation explicitly. Our model maximizes the advantage of the light-speed natural of optical computing by implementing larger convolution kernels (e.g., 7$\times$7, 11$\times$11). The proposed LMNN yields larger reception fields, without sacrificing low computational latency and low energy consumption. Furthermore, the aforementioned physical limitations of LMNNs are explicitly addressed via optimized digital modeling. We evaluate our model on image classification tasks using two public datasets: FashionMNIST~\cite{xiao2017fashion} and STL-10~\cite{coates2011analysis}. The proposed LMNN achieved superior classification accuracy as compared with the SOTA MNN and model re-parametrization methods. Overall, the system's contributions can be summarized in four-fold:

\begin{figure}
\begin{center}
\includegraphics[width=0.45\linewidth]{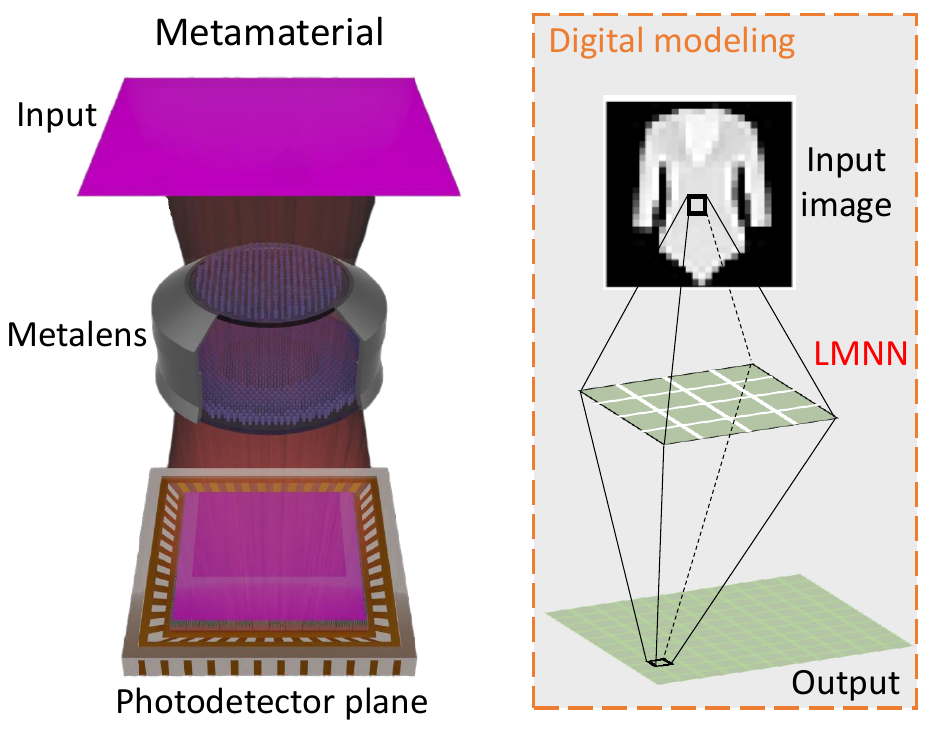}
\end{center}
   \caption{This study provides a digital modeling platform for designing and optimizing a metamaterial neural network (MNN). The proposed large kernel metamaterial neural network (LMNN) is able to maximize the performance of an MNN without introducing extra computational complexity during the inference stage.}
\label{fig:fig1}
\end{figure}

\begin{itemize}
\item We propose the large convolution kernel design for an LMNN to achieve a larger reception field, lower computational latency, and less energy consumption. 
\item We introduce the model re-parameterization and multi-layer compression mechanism to compress the multi-layer multi-branch design to a single layer for the LMNN implementation. This maximizes the model capacity without introducing any extra burden during the optical inference stage.
\item The physical limitations of LMNNs (e.g., limited kernel size, channel number, precision, noise, non-negative restriction, and bandwidth) are explicitly addressed via optimized digital modeling.
\item We implemented a one-layer LMNN with real physical metamaterial fabrication to demonstrate the feasibility of our hybrid design. 
\end{itemize}

\begin{figure*}
\begin{center}
\includegraphics[width=0.9\linewidth]{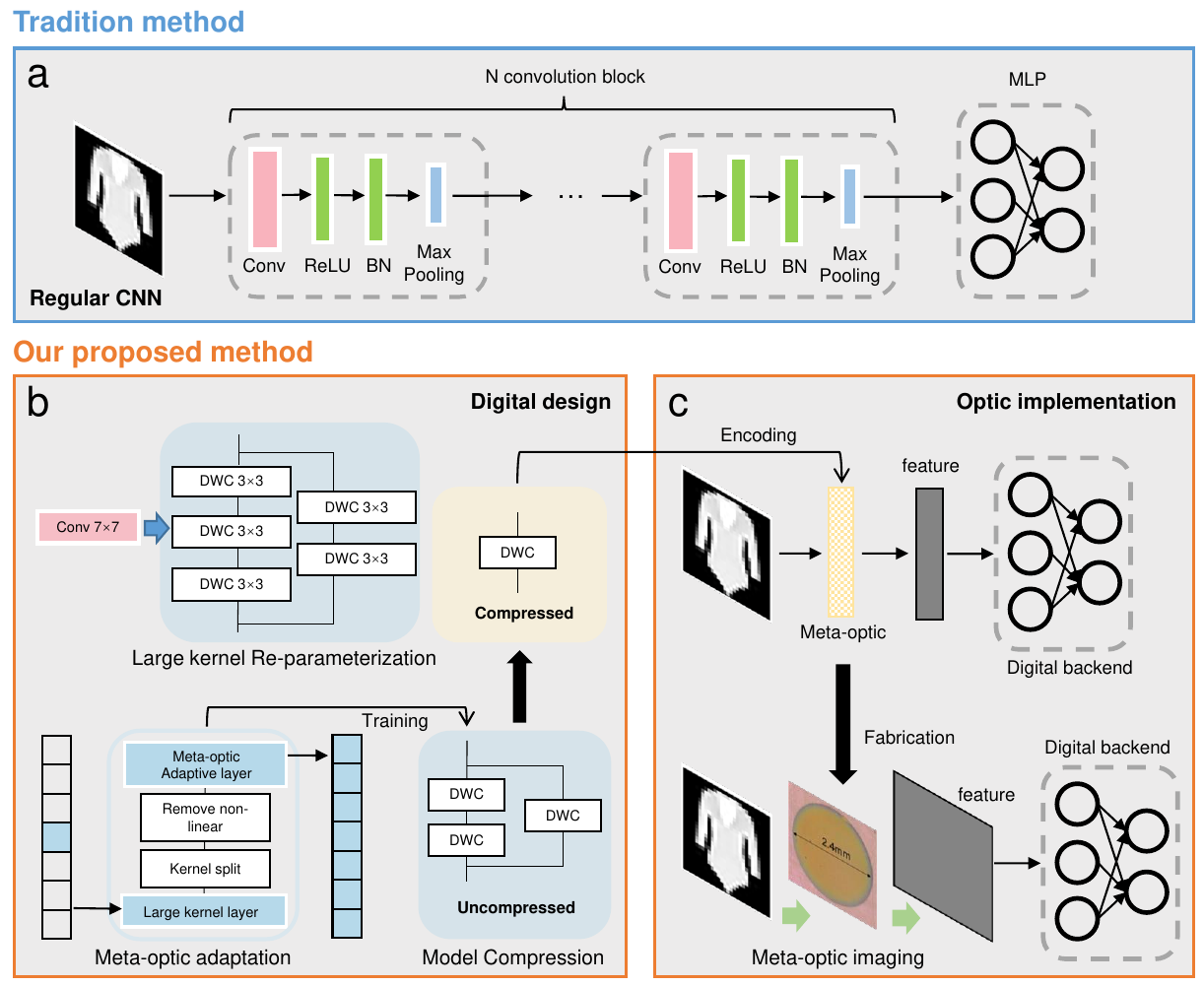}
\end{center}
   \caption{The upper panel (a) shows the conventional CNN model on the image classification task. The lower panels (b) present our proposed LMNN method with digital design and optic implementation. The large kernel re-parameterization efficiently achieves a large receptive field with a multi-branch multi-layer structure. Physical constraints are modeled via the meta-optic adaptation. The multi-branch multi-layer model is further compressed to a single-layer LMNN. (c) The digital design is fabricated as a real meta-optic device for inference.}
\label{fig:structure}
\end{figure*}

\section{Related work}
\subsection{Models with large kernel convolution}

For a decade, a common practice in choosing optimal kernel size in convolution is to leverage 3$\times$3 kernel. In recent years, more attention has been put into a larger kernel design. The Inception network proposes an early design of adapting large kernels for vision recognition tasks~\cite{chollet2017xception}. After developing several variations~\cite{szegedy2015going, szegedy2016rethinking}, large kernel models became less popular. Global Convolution Networks (GCNs)~\cite{peng2017large} employ the large kernel idea by utilizing 1$\times$K followed by K$\times$1 to achieve improvement in model performance for semantic segmentation. 

Current limitations of leveraging large kernel convolution kernel can be divided in two aspects: (1) scaling up the kernel sizes lead to the degradation of model performance, and (2) its high computational complexity. According to the Local Relation Networks (LRNet)~\cite{hu2019local}, the spatial aggregation mechanism with dynamic convolution is used to substitute traditional convolution operation. As compared with the traditional 3$\times$3 kernels, the LRNet~\cite{hu2019local} leverage 7$\times$7 convolution to improve model performance by a significant margin. However, the performance become saturated by scaling up the kernel size to 9$\times$9. Similar to RepLKNet ~\cite{ding2022scaling}, scaling up the convolution kernel size to 31$\times$31 without prior structural knowledge demonstrates the decrease of model performances. 

\subsection{Model compression and re-parameterization}
Though many complicated ConvNets~\cite{iandola2014densenet, huang2018condensenet} deliver higher accuracy than more simple ones, the drawbacks are significant.
1) The complicated multi-branch designs (e.g., residual addition in ResNet~\cite{he2016deep} and branch-concatenation in Inception~\cite{szegedy2015going})
make the model difficult to implement and customize, and slow
down the inference and reduce memory utilization. 2)
Some components (e.g., depthwise convolution in Xception~\cite{chollet2017xception}
and MobileNets~\cite{howard2017mobilenets}, and channel shuffle in ShuffleNets~\cite{zhang2018shufflenet}) increase memory access costs and lack support for various devices.

Model compression~\cite{cheng2018model} aims to reduce the model size and computational complexity~\cite{vanhoucke2011improving,chen2015compressing} while maintaining their performance including pruning and quantization. Pruning has been widely used to compress deep learning models by removing the unnecessary or redundant parameters from a neural network without affecting its accuracy~\cite{srinivas2015data,han2015learning,he2017channel}. Quantization has two categories: Quantization-Aware Training (QAT)~\cite{gong2014compressing,wu2016quantized} and Post-Training Quantization (PTQ). QAT applies quantization operation in the training stage. In contrast, PTQ takes a full precision network for training and quantized it in the post stage~\cite{liu2021post,fang2020post,li2021brecq}. 

\subsection{Optical neural network}
The Optical neural network has high band width~\cite{zhou1994acoustic}, uses light instead of electrical signals to perform matrix multiplications~\cite{larger2012photonic,duport2012all} which can be much faster and more energy-efficient than traditional digital neural networks. Most optical neural networks (ONN) use a hybrid model structure: implement linear computation with optic device and non-linear operation digitally~\cite{jutamulia1996overview, paquot2012optoelectronic, woods2012photonic, hughes2018training}. Besides the use of optical devices, ONN has been implemented on nanophotonic circuits~\cite{fang2015nanoplasmonic, shen2017deep} and light-wave linear diffraction~\cite{ovchinnikov1999diffraction, lin2018all} to improve model efficiency. For the non-linear computation, \cite{george2018electrooptic, miscuglio2018all} have proposed implementing the non-linear operation with the optic device on ONN.

\section{Method}
\textbf{Problem statement.} The goal of this study is to develop a new digital learning scheme to maximize the learning capacity of MNN while modeling the physical restrictions of meta-optic. With the proposed LMNN, the computation cost of the convolutional front-end can be offloaded into fabricated optical hardware, so as to get optimal energy and speed efficiency under current fabrication limitations. We adapt our innovations with four aspects: (1) large kernel re-parameterization, (2) meta-optic adaptation, and (3) model compression.

\begin{figure}
\begin{center}
\includegraphics[width=0.80\linewidth]{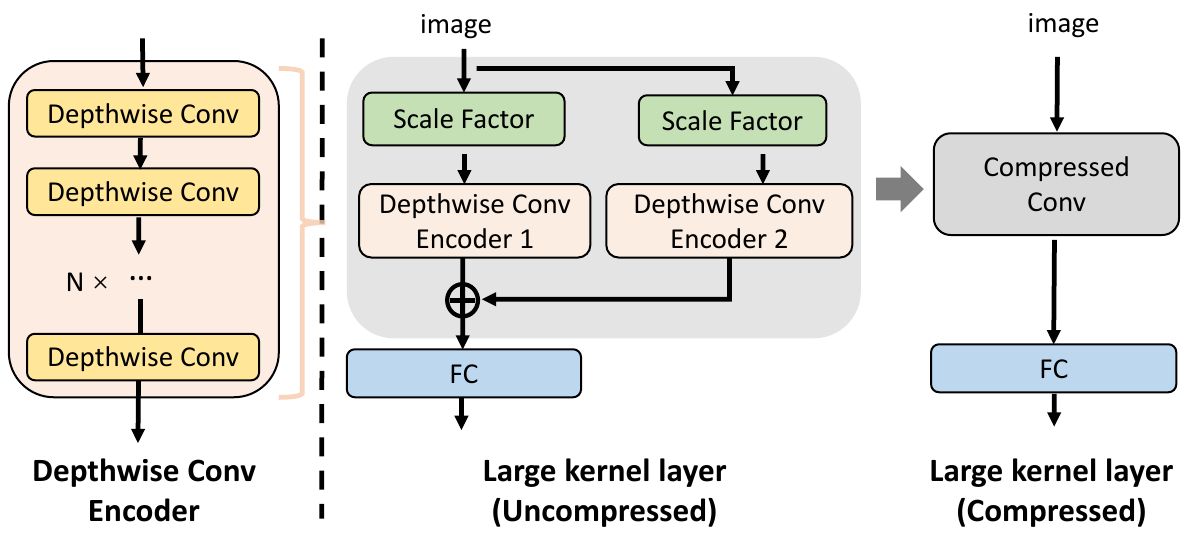}
\end{center}
   \caption{An overview of our proposed large convolution kernel block with re-parameterization is presented. Learnable scaling factors are employed to mimic the scaling function of batch normalization. In the inference stage, the block can be converted to a single convolution layer.}
\label{fig:fig3}
 \end{figure}

\subsection{Large kernel re-parameterization}
To tackle the limitation of only fabricating the first layer only in CNNs, we need to maximize the performance of the first layer, while it is feasible to adapt the fabrication processing. With the significant progress in Vision Transformers (ViTs), the key contribution for the performance gained is largely credited to the large effective receptive field, which can be generated similarly by the depthwise convolution with large kernel sizes in CNNs. Therefore, we explore the feasibility of adapting large kernel convolution in 1) single-branch and 2) multi-branch setting.\\
\textbf{Single Branch Design.} Inspired by ~\cite{ding2022scaling}, a large depthwise convolution kernel is equivalently to have the same receptive fields to a stack of small kernels. With the intrinsic structure of depthwise convolution, such a stack of kernel weights can be compressed into a single operator. It is thus essential for the LMNN to maximize the model performance via a relatively simple meta-optic design, with a single compressed convolution layer. The compressed design further introduces fewer model FLOPs in the model inference stage. For conventional convolution operation, the convolution weight matrix $\textbf{W} \in \mathbb{R}^{C_{i} \times C_{o} \times K_{h} \times K_{w}}$. The $C_{i}$ and $C_{o}$ are input channel and output channel of the convolution layer. $K_{h}$ and $K_{w}$ are height and width of convolution kernel. Denote we have an input patch $x$ in size of $H \times W$ and the output is $y$, we have conventional convolution as equation \ref{eq:tradition_conv}.
\begin{equation} \label{eq:tradition_conv}
	y = W * x
\end{equation}
where $ y = \sum_{p=0}^{n_{i}} W_{p}*x_{p} $, $*$ represents convolution between matrices. For the input $x$, the computation time complexity will be $O(H \times W \times C_{i} \times C_{o} \times K_{h} \times K_{w})$. 
For the depthwise convolution model, channels $C_{i}$ in the convolution layer are separated along with the input data channels of $x$. The depthwise convolution follows the equation. \ref{eq:depth_conv}
\begin{equation} \label{eq:depth_conv}
	y_{i}' = W_{i} * x_{i}
\end{equation}
where $y_{i}$ is the $i$th channel of output $y$, $W_{i}$ and $x_{i}$ are the $i$th channel from Convolution weight $W$ and input data $x$, respectively. The time complexity is $O(H \times W \times C_{i} \times K_{h} \times K_{w})$. Normally, the input channel number equals to the output channel number. We can infer the theoretical speed-up ratio $r$ on model FLOPs between convention convolution and depthwise convolution following the equation.\ref{eq:speed_ratio}
\begin{equation} \label{eq:speed_ratio}
	r = \frac{O(H \times W \times C_{i} \times C_{o} \times K_{h} \times K_{w})}{O(H \times W \times C_{i} \times K_{h} \times K_{w})} = O(C_{i})
\end{equation}
where $C_{i}$ is the channel number of the convolution layer. Depthwise convolution has $C_{i} = C_{o} = C$. The depthwise convolution operation saves more FLOPs when the channel number is large compared with convention convolution.\\
\textbf{Multi-branch design.} Inspired by RepVGG ~\cite{ding2021repvgg} and RepLKNet ~\cite{ding2022scaling}, the multi-branch design demonstrates the feasibility of adapting large kernel convolutions (e.g., $31\times31$) with optimal convergence using a small kernel convolution in parallel. The addition of the encoder output enhances the large kernel convolution in the locality. According to the properties of convolution operation, the abstracted feature map from the parallel convolution path can be overlapped by learning different features. By using different convolution kernel sizes, the features from different scales of view are abstracted simultaneously. 

We denote that output $y'$ and input patch $x$ use a two-branch convolution block $W$. 
\begin{equation} \label{eq:reparameter}
	y_{i}' = W_{1} * x + W_{2} * x
\end{equation}
where $W_{1}$ and $W_{2}$ is two different convolution layer with different kernel size. For multiple parallel paths, the $N$-branch convolution can be generated as equation \ref{eq:reparameter-N}.
\begin{equation} \label{eq:reparameter-N}
	y = \sum_{q=0}^{N} W_{q}*x
\end{equation}
According to the equation \ref{eq:reparameter-N}, output $y$ has the feature map from multiple scales of views. The overlap of convolution output from different scale redistributes the feature map which is proved by~\cite{ding2022scaling} to have better performance.

\subsection{Meta-optic adaptation}
As to integrate the large kernel convolution design into meta-optic devices, we need to consider and model the physical restrictions explicitly in our model design, beyond the conventional digital training (Fig.~\ref{fig:structure}). First, the weight in convolution kernel should be positive for fabrication. Second, the convolution layer that substitutes by metalens should be the first layer of the model. Third, metalens can only take single-channel images. Thus, all RGB images are transferred to grayscale images. Fourth, due to the optic implementation purpose, the size of the convolution kernel is limited. Last, the channel number of the convolution layer is limited by the size of the optic device capacity.

\begin{figure}
\begin{center}
\includegraphics[width=0.95\linewidth]{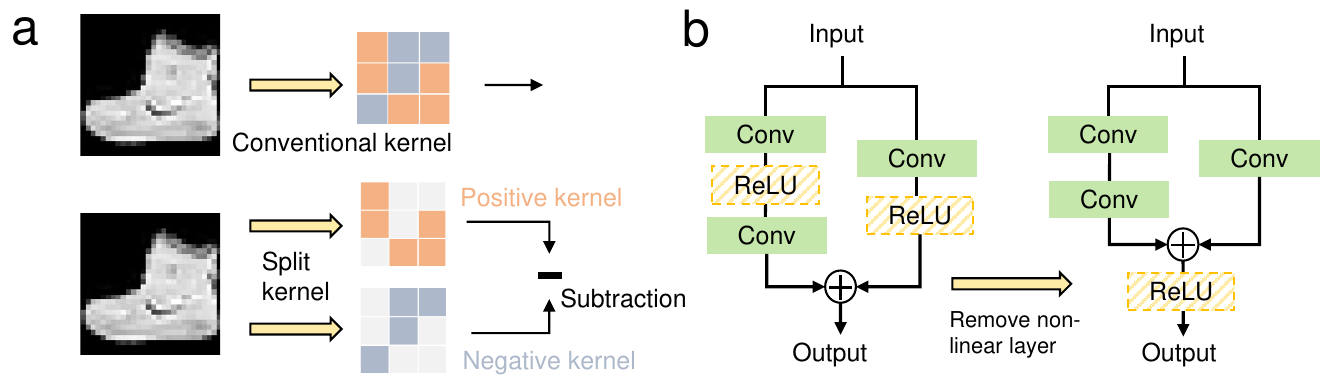}
\end{center}
   \caption{Adaptation for meta-optic implementation. (a) To implement the kernel with negative weight, we split the kernel into the positive kernel and negative kernel and subtract from their feature map. (b) The non-linear layer needs to be removed from the parallel convolution path.}
\label{fig:solution}
 \end{figure}

\textbf{Split kernel.} To keep the model convolution kernel weight positive for the optic device implementation, we split the convolution kernel into two part: positive weight and negative weight. As shown in Fig.\ref{fig:solution}, the final convolution kernel results are the subtraction of the two feature map from the positive and negative convolution kernel respectively.

\textbf{Remove non-linear layer.}   
In traditional convolution operation, non-linear layer is typically added between the convolution layers. The non-linear layer, including batch normalization and activation layers (eg. ReLU) introduce the non-linear transformation to the model. However, they are not able to be implemented with our meta-optic device. As shown in Fig.\ref{fig:solution}, the non-linear layers are removed from the parallel convolution branch and connected behind the large kernel convolution layer.

\textbf{Non-negative weight in optic kernel}
In traditional deep learning model, both positive weights and negative weights are stored. The meta-optic model implementation can only take positive kernel weights. Adaptation methods are applied to convolution model training to constrain model weight to a positive value. Four methods are introduced in model training: square of trigonometric functions, mask out the negative value, add non-negative loss, and our proposed kernel split. 

Square of trigonometric function: instead of directly updating the weight, we define weight as equation \ref{eq:sin2}. The weight $W_{i}$ keeps positive and in range $[0, 1]$ whatever the value of $\theta$.

\begin{equation} \label{eq:sin2}
	W_{i} = Sin^{2}(\theta_{i})
\end{equation}

Mask out the negative value: in the training process, the weight smaller than 0 is assigned as 0 manually after each iteration update.

Add non-negative loss: to maintain the model weight positive, a non-negative weight loss is add to the loss function, which is defined as equation~\ref{eq:non_neg_loss}.

\begin{equation} \label{eq:non_neg_loss}
	loss = \sum (model.weight < 0)
\end{equation}

\textbf{Bandwidth and precision.} Due to the accuracy of the current fabrication of meta-optic, the optical inference might lose precision. As a result, the model bandwidth and weight precision should also be modeled during the training process. For example, PyTorch has a default 32-bit precision, which is not feasible for the LMNN. Thus, the quantize is employed to simulate the model performance when all digital neural networks are implemented with optic devices. Taking the noise in optic implementation into consideration, which will affect the model weights precision, we add the Gaussian noise to the digital convolution weight. 

\subsection{Model compression}
The stacked depthwise convolution and re-parameterization can potentially improve the model performance by learning with variance. The multilayer structure can be regarded as multiple stacked depthwise convolution layers which make the model deeper. The multi-branch structure will make the model wider. It is obvious the designed model is in a complex structure. To save image processing time in the inference stage, the multiplayer structure can be squeezed into a single layer. In this paper, we only explore the squeezed convolution layer. To get the equivalent squeezed layer, a non-linear component should be eliminated. The non-linear layers such as activation function and batch normalization are moved out of our squeezed block. The stacked convolution kernel follows the equation \ref{eq:stacked_depth}. 

\begin{equation}
\begin{aligned}
\label{eq:stacked_depth}
	y = & (W_{N} * (W_{N-1} * \dots\dots\dots (W_{2} * W_{1}))) * x \\
       = & W^{*} * x
\end{aligned}
\end{equation}

\begin{equation} \label{eq:eqa_w}
	W^{*} = (W_{N} * (W_{N-1} * \dots\dots\dots (W_{2} * W_{1})))
\end{equation}

$W^{*}$ is the equivalent weight to the stacked setting in equation \ref{eq:eqa_w}.
As the number of stacked convolution layers increases, the equivalent convolution kernel is larger. The equivalent kernel size k and the number of stacked $3\times3$ convolution layer n follow equation~\ref{eq:kernel_size}.
\begin{equation} \label{eq:kernel_size}
	k = 2 \times n +1
\end{equation}

For example, two 3$\times$3 convolution kernels are equivalent to a 5$\times$5 convolution kernel. The multi-branch convolution layer can be compressed as shown in Fig~\ref{fig:fig3}.

Since the convolution kernel value from the different parallel branches is equivalent to a single kernel by overlapping kernel, a multi-parallel convolution branch can be compressed into a single path.

\section{Data and experimental design}

\begin{figure*}
\begin{center}
\includegraphics[width=0.8 \linewidth]{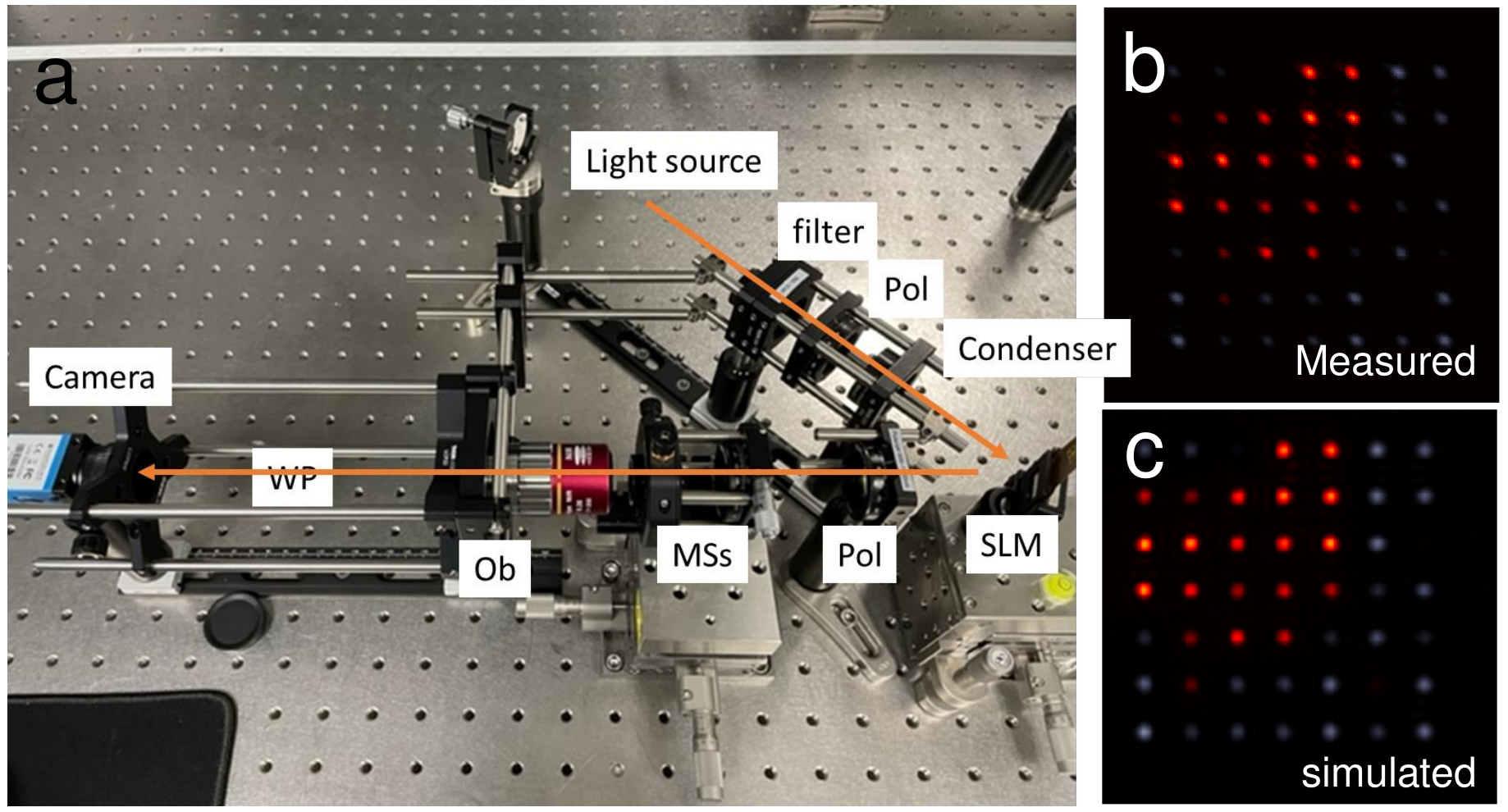}
\end{center}
   \caption{The meta-optic devices simulation and implementation platform. (a) Optic system for meta-optic lens test. The components in the figure are: Light source: Tungsten Lamp; filter: Wavelength filter; Pol: Polarizer; SLM: Spatial light modulator; Condenser: Lens to focus light on the SLM; MSs: Metasurfaces; Ob: Objective lens. (b) Measured meta-optic kernel weight point spread function, used for optical convolution with the imaged object. (c) Theoretical meta-optic kernel weight point spread function by simulation.}
\label{fig:metaoptic}
 \end{figure*}

\subsection{Data description}
Two public datasets, FashionMNIST~\cite{xiao2017fashion} and STL10~\cite{coates2011analysis}, were employed to evaluate the performance of the proposed method on image classification tasks. For the FashionMNIST dataset, we employed 60,000 images for training and 10,000 images for testing. The images were grayscale images in the size of 28 $\times$ 28. FashionMNIST was inspired by the MNIST dataset, which classified clothing images rather than digits. We employed STL-10 as another cohort with a larger input image size (96$\times$96). In our experiments, the RGB images in STL-10 were transferred to grayscale images due to the physical limitation in the LMNN.

\subsection{Large kernel re-parameterization}
We proposed the large re-parameterized convolution kernel design in our LMNN network to maximize the computational performance of the precious single metamaterial layer by (1) taking advantage of high-speed light computation, and (2) overcoming the physical limitations in an MNN implementation. 

 To evaluate the large re-parameterized convolution kernel on FashionMNIST, we constructed a naive model that consisted of a large re-parameterized convolution kernel block, a single fully connected layer, as well as non-linear components (ReLU activation, batch normalization, and the softmax function). Different re-parameterization model structures were evaluated. To demonstrate the impacts of the size, the kernel was tested from 3$\times$3 to 31$\times$31. Besides the kernel size, we evaluated multiple numbers of parallel branches: from a single path to four paths.
 

\subsection{Meta-optic model adaptation}

The performance of the LMNN is fundamentally limited by physical restrictions. We provide the model simulation by modeling optic system limitations. In regards to the model limitations, the convolution kernel is implemented with optical devices that can only have limited channels. To include the meta-optic devices in our network, the layer that is to be substituted should be the first layer of our model. The following model structure can be designed digitally. To validate model design on different sizes, deep neural networks with multiple convolution layers are implemented. 

To simulate the noise in real meta-optic fabrication, we add random noise following the Gaussian distribution. To test the impact of noise level, we simulate the noise amplitude range from 0.05 to 0.2. Considering the meta-optic implementation on the whole model for further research, we quantize the model weight in 8-bit instead of the default 32-bit.

In order to evaluate the non-negative weight effect, three methods are evaluated to constrain the model weight positive. 'Sin' in Fig.~\ref{fig:abla}(a) means weights are defined by square of sin function. 'Mask out' is to eliminate the negative weight by screening out. Loss function is also used to define model with positive weights, which results is shown in Fig.~\ref{fig:abla}(a) as 'Non-neg' loss.

The large kernel convolution design is validated on fabricated meta-optic devices. Based on the well-trained digital convolution kernel weight, meta-optic lenses are implemented and tested in real optic systems shown in Fig.~\ref{fig:metaoptic}.

\subsection{Model compression efficiency}
Through model compression, the model in the inference stage alleviates the computation load with lighter weights. The fabricated convolution kernel by a meta-optic lens with the digital backend is assembled as the hybrid model. We test the model's inference time by feeding the same image and recording the model's processing time.

To test the optimal LMNN structure under the meta-optic fabrication limitation, the combination of layer numbers from one to five and channel numbers from nine to twenty. The model digital computation load (FLOPs) and  the ratio of meta-optic is computed to find the model structure achieves optimal efficiency.

\section{Result}

In this section, we first evaluate our proposed large kernel network with a simple model structure, using the FashionMNIST dataset and STL-10 dataset. We then evaluate the large kernel capability on complex convolution neural networks with the same dataset.

\begin{table}
\centering
\caption{Large re-parameterized convolution experiment results}
\label{tab:all_result}
\begin{tabular}{lcccc}
    \hline
        & \multicolumn{2}{c}{FashionMNIST}   &  \multicolumn{2}{c}{STL-10}  \\
        &  Model Conv    &  Test     &  Model Conv    & Test   \\
    \hline
    Naive model   &   3$\times$3    &  0.8495   &   3$\times$3    &  0.4500      \\
    RepLKNet \cite{ding2022scaling} &  7$\times$7   &  0.9015  &  7$\times$7   &  0.4993  \\
                                    &       &      &  11$\times$11    &  0.5241   \\
    RepVGG~\cite{ding2021repvgg}    & 7+5+3  &   0.9081   &  7+5+3    &  0.5341      \\
                                    &       &      &  11+9+7   &  0.5650      \\
    Depthwise conv~\cite{chollet2017xception}    &   3 dwc    &  0.9084  & 3 dwc   &  0.5509  \\
                                    &       &      & 5 dwc   & 0.5935   \\ 
    LMNN (Ours)   &  3 dwc + 2 dwc + 1 dwc   &  \bf{0.9115}  &  5 dwc + 3 dwc + 1 dwc   &  \bf{0.6120}  \\
    \hline

\end{tabular}
\footnotetext[1]{'dwc' refer to the depthwise convolution layer, convolution kernel size is $3\times3$}
\end{table}

\subsection{Large re-parameterized convolution performance}
We evaluate the large re-parameterized convolution model on two datasets: FashionMNIST and STL-10. As shown in Table.~\ref{tab:all_result}, the naive model with 7$\times$7 convolution kernels has demonstrated better performance than that with $3\times 3$. With structural reparameterization,  the model prediction accuracy further improves. Meanwhile, the model implemented with a depthwise convolution layer outperformed the baselines with both small and large convolution kernels. 

Our large kernel model is evaluated on the STL10 dataset with a larger image size (96$\times$96). As compared with performance on FashionMNIST (image size 30$\times$30), the large kernel convolution model reveals greater improvements, as shown in Table.~\ref{tab:all_result}. The model with 11$\times$11 kernel size has better accuracy (0.5341) compared with that of using 3$\times$3 and 7$\times$7. By integrating the depthwise convolution design, the model performance boosts from 0.5241 to 0.5935. Our proposed large kernel block outperforms all SOTA approaches and achieves the best accuracy of 0.6015 with teacher model supervised training.

\begin{figure}[h]
\centering
\includegraphics[width=0.55 \linewidth]{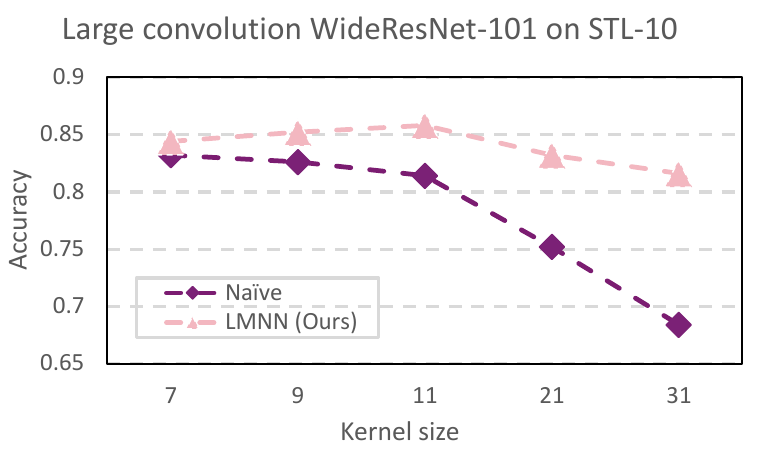}
\caption{Large convolution WideResNet-101 model performance on STL-10.}
\label{fig:widres101}
\end{figure}

To further validate our large kernel with depthwise convolution design, we conduct experiments on more sophisticated models by replacing all convolution layers with the large re-parameterized convolution layers. Briefly, WideResNet-101 is used as complex model backbone \cite{kabir2022spinalnet}. Model performance is shown in Table.~\ref{fig:widres101}. By substituting the first convolution layer with a larger kernel size, the model performance improves from 0.94 to 0.96 with utilizing larger images (256$\times$256 RGB).

\begin{figure*}[t]
\begin{center}
\includegraphics[width=1\linewidth]{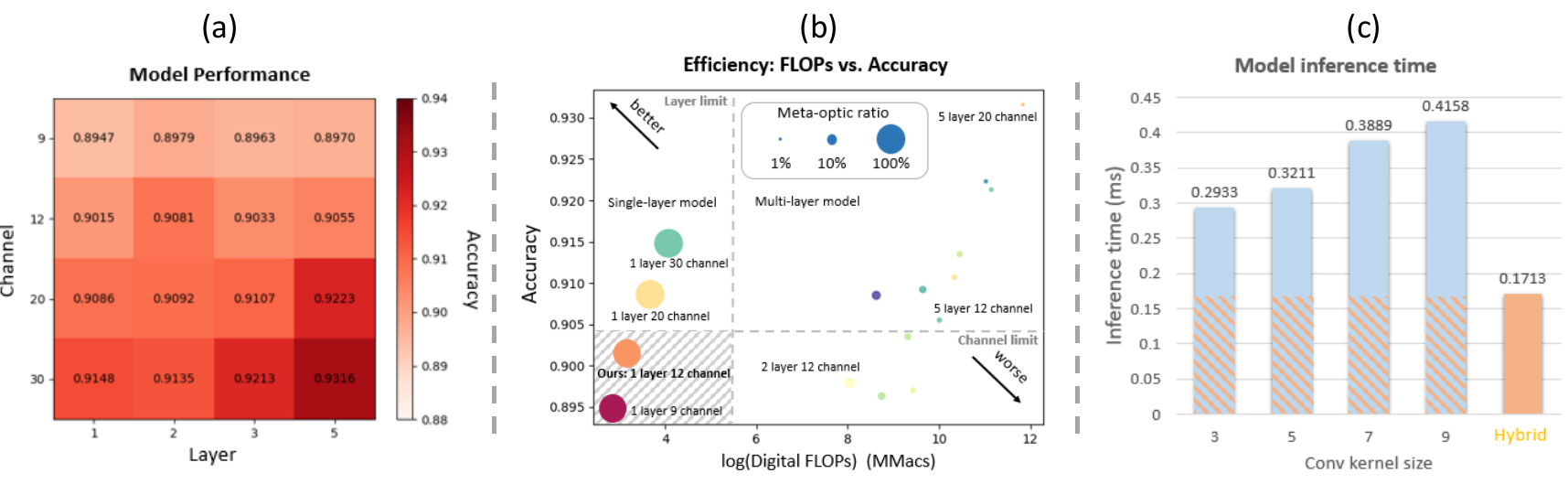}
\end{center}
   \caption{(a) The large re-parameterized convolution model performance with different layer numbers and channel numbers. (b) Large convolution kernel efficiency evaluation. The circle in different colors shows different convolution layer structures. The shadow area is the model structure that can be fabricated. The circle area shows the FLOPs ratio of the layer implemented by meta-optic material. x-axis is the model FLOPs except the layer to be fabricated. (c) Model inference time between the digital model and hybrid model. The orange shadow area in the figure shows the time used by the model rest part except the large convolution kernel layer.}
\label{fig:cm}
 \end{figure*}

 \subsection{Performance of model adaptation}
To validate our large kernel design on the real metasurface fabrication model shown in Fig.~\ref{fig:metaoptic}, we implement a model trained on FashionMNIST with a large kernel design, utilizing a digital design for comparison. The digital convolution layer has 12 channels 7$\times$7 convolution kernel which is the optimal kernel design under the current meta-optic implement limitation. As shown in Table.\ref{tab:optic_implement}, the Metamaterial Neural Network demonstrates excellent consistency with the theoretical performance of a digital neural network. 

\begin{table}[h]
\caption{Metasurface fabrication}
\begin{center}
  \centering
  \begin{tabular}{cc}
    \hline
    Method  &  Test  \\
    \hline
    Digital Neural Network (DNN)   &  0.9015    \\
    Large Kernel MNN (LMNN)   &  0.8760      \\
    \hline
  \end{tabular}
  \label{tab:optic_implement}
\end{center}
\end{table}

Due to the meta-optic implementation limitation, four adaptation methods are applied to constrain kernel weights to positive. According to the model performance, our proposed kernel split method shows superior performance over the common training strategies.

\subsection{Ablation studies}
\begin{figure}[h]
\begin{center}
\includegraphics[width=0.95 \linewidth]{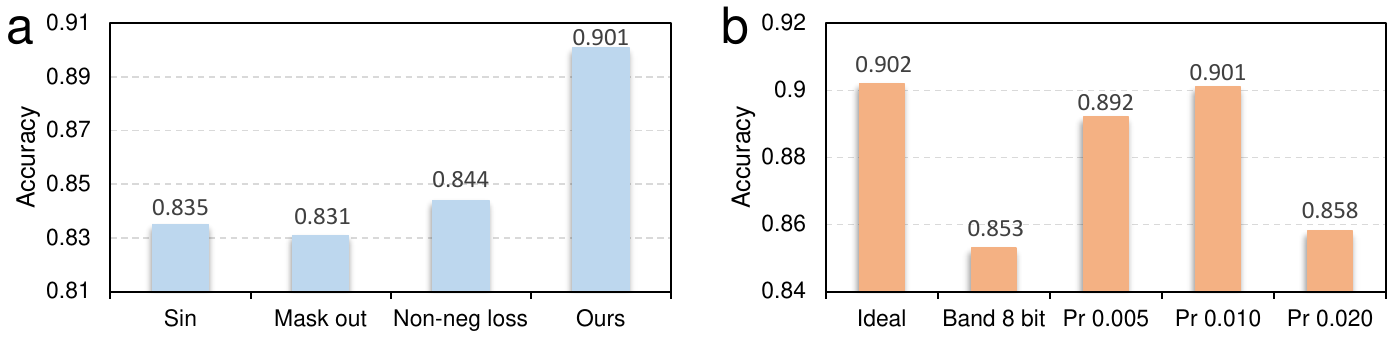}
\end{center}
   \caption{Plot of ablation study on LMNN. (a) Evaluating non-negative weight effect on model performance. (b) Measuring the effect of model bandwidth and weight precision effect on model prediction accuracy. 'Pr' in Figure (b) means precision.}
\label{fig:abla}
 \end{figure}

To validate our model bandwidth and weight precision limit simulation, the results of the experiment are shown in Table~\ref{fig:abla}.

To evaluate the upper bound performance on FashionMNIST,
a deep model structure is implemented and tested on FashionMNIST. The number of convolution layers in our model range from 1 to 5, and the channel number ranges from 9 to 30. The model performance is shown in Fig.~\ref{fig:cm} (a). The model with more parameters shows a higher accuracy. Regardless of the meta-optic fabrication limitation, the meta-optic hybrid model achieves better performance.

\subsection{LMNN efficiency and speed evaluation}
To evaluate the model on both speed and computation load, we compute the model FLOPs except the large convolution layer and the FLOPs ratio of the layer implemented by meta-optic material. The model performance with different structure is shown in Fig.~\ref{fig:cm} (b). The optimal model structure should at top left corner in shadow area. As shown, the model with 1 large re-parameterized convolution layer and 12 channels is the optimal structure. To show the speed advantage of our LMNN, the model inference time is recorded. From Fig.~\ref{fig:cm} (c), the hybrid model shows a speed twice fast as compared to the digital convolution model.

\section{Discussion}

In this work, we present a convolution block with a large kernel design that generates larger receptive fields to maximize the digital capacity of LMNN. To validate the large kernel convolution design, we further applied the block to a complex model such as WideResNet-101. From the experiments, two important components contribute to the improvement of large kernel design from traditional 3$\times$3 kernel size. First, the larger convolution kernel can get larger receptive fields. According to the target image size, the convolution kernel size is not the larger the better. For the FashionMNIST in size of 30$\times$30, 7$\times$7 is the best kernel size. For the images from STL-10 dataset in size of 96$\times$96, 11$\times$11 kernel performed the best. Another interesting point is the stacked depthwise convolution layers have equivalent computing operations to the single convolution layer with a larger kernel size. The multi-layer depthwise convolution and multi-branch structure expand the model capacity without parameter increase. 

Since the large convolution kernel achieved superior performance on image classification task, more computer vision tasks have the improvement potential. For the image segmentation task, it can be regarded as a pixel-level classification problem. The large convolution design can be applied on segmentation tasks. Object detection can be another choice for large convolution kernel application. Different size of convolution kernel provides multiple field of views. The views from multiple scale can abstract representation with more spatial information.

\section{Conclusion}
In this paper, we present a large-kernel convolution block and implementation on meta-optic lens. The evaluation is conducted on FashionMNIST and STL-10 dataset. We revisit the model design of a metamaterial neural network by modeling the physical restrictions explicitly. Our proposed LMNN outperformed the benchmarks on both datasets. Moreover, the computational latency is reduced by using light-speed optical convolution.

\section*{Acknowledgment}

HZ BTS and JGV acknowledge support from DARPA under contract HR001118C0015, NAVAIR under contract N6893622C0030 and ONR under contract N000142112468. YH and QL acknowledge support from NIH under contract R01DK135597. YH is the corresponding author.

\ifCLASSOPTIONcaptionsoff
  \newpage
\fi



\bibliographystyle{IEEEtran}
%



\bibliography{jist_lmnn}

\begin{thebibliography}{10}
\providecommand{\url}[1]{#1}
\csname url@samestyle\endcsname
\providecommand{\newblock}{\relax}
\providecommand{\bibinfo}[2]{#2}
\providecommand{\BIBentrySTDinterwordspacing}{\spaceskip=0pt\relax}
\providecommand{\BIBentryALTinterwordstretchfactor}{4}
\providecommand{\BIBentryALTinterwordspacing}{\spaceskip=\fontdimen2\font plus
\BIBentryALTinterwordstretchfactor\fontdimen3\font minus
  \fontdimen4\font\relax}
\providecommand{\BIBforeignlanguage}[2]{{%
\expandafter\ifx\csname l@#1\endcsname\relax
\typeout{** WARNING: IEEEtran.bst: No hyphenation pattern has been}%
\typeout{** loaded for the language `#1'. Using the pattern for}%
\typeout{** the default language instead.}%
\else
\language=\csname l@#1\endcsname
\fi
#2}}
\providecommand{\BIBdecl}{\relax}
\BIBdecl

\bibitem{lecun1989backpropagation}
Y.~LeCun, B.~Boser, J.~S. Denker, D.~Henderson, R.~E. Howard, W.~Hubbard, and
  L.~D. Jackel, ``Backpropagation applied to handwritten zip code
  recognition,'' \emph{Neural computation}, vol.~1, no.~4, pp. 541--551, 1989.

\bibitem{krizhevsky2017imagenet}
A.~Krizhevsky, I.~Sutskever, and G.~E. Hinton, ``Imagenet classification with
  deep convolutional neural networks,'' \emph{Communications of the ACM},
  vol.~60, no.~6, pp. 84--90, 2017.

\bibitem{li2014medical}
Q.~Li, W.~Cai, X.~Wang, Y.~Zhou, D.~D. Feng, and M.~Chen, ``Medical image
  classification with convolutional neural network,'' in \emph{2014 13th
  international conference on control automation robotics \& vision
  (ICARCV)}.\hskip 1em plus 0.5em minus 0.4em\relax IEEE, 2014, pp. 844--848.

\bibitem{jha2020doubleu}
D.~Jha, M.~A. Riegler, D.~Johansen, P.~Halvorsen, and H.~D. Johansen,
  ``Doubleu-net: A deep convolutional neural network for medical image
  segmentation,'' in \emph{2020 IEEE 33rd International symposium on
  computer-based medical systems (CBMS)}.\hskip 1em plus 0.5em minus
  0.4em\relax IEEE, 2020, pp. 558--564.

\bibitem{ronneberger2015u}
O.~Ronneberger, P.~Fischer, and T.~Brox, ``U-net: Convolutional networks for
  biomedical image segmentation,'' in \emph{Medical Image Computing and
  Computer-Assisted Intervention--MICCAI 2015: 18th International Conference,
  Munich, Germany, October 5-9, 2015, Proceedings, Part III 18}.\hskip 1em plus
  0.5em minus 0.4em\relax Springer, 2015, pp. 234--241.

\bibitem{chauhan2018convolutional}
R.~Chauhan, K.~K. Ghanshala, and R.~Joshi, ``Convolutional neural network (cnn)
  for image detection and recognition,'' in \emph{2018 first international
  conference on secure cyber computing and communication (ICSCCC)}.\hskip 1em
  plus 0.5em minus 0.4em\relax IEEE, 2018, pp. 278--282.

\bibitem{redmon2016you}
J.~Redmon, S.~Divvala, R.~Girshick, and A.~Farhadi, ``You only look once:
  Unified, real-time object detection,'' in \emph{Proceedings of the IEEE
  conference on computer vision and pattern recognition}, 2016, pp. 779--788.

\bibitem{liu2021swin}
Z.~Liu, Y.~Lin, Y.~Cao, H.~Hu, Y.~Wei, Z.~Zhang, S.~Lin, and B.~Guo, ``Swin
  transformer: Hierarchical vision transformer using shifted windows,'' in
  \emph{Proceedings of the IEEE/CVF International Conference on Computer
  Vision}, 2021, pp. 10\,012--10\,022.

\bibitem{wang2021pyramid}
W.~Wang, E.~Xie, X.~Li, D.-P. Fan, K.~Song, D.~Liang, T.~Lu, P.~Luo, and
  L.~Shao, ``Pyramid vision transformer: A versatile backbone for dense
  prediction without convolutions,'' in \emph{Proceedings of the IEEE/CVF
  international conference on computer vision}, 2021, pp. 568--578.

\bibitem{liu2022convnet}
Z.~Liu, H.~Mao, C.-Y. Wu, C.~Feichtenhofer, T.~Darrell, and S.~Xie, ``A convnet
  for the 2020s,'' in \emph{Proceedings of the IEEE/CVF Conference on Computer
  Vision and Pattern Recognition}, 2022, pp. 11\,976--11\,986.

\bibitem{ding2022scaling}
X.~Ding, X.~Zhang, J.~Han, and G.~Ding, ``Scaling up your kernels to 31x31:
  Revisiting large kernel design in cnns,'' in \emph{Proceedings of the
  IEEE/CVF Conference on Computer Vision and Pattern Recognition}, 2022, pp.
  11\,963--11\,975.

\bibitem{liu2022more}
S.~Liu, T.~Chen, X.~Chen, X.~Chen, Q.~Xiao, B.~Wu, M.~Pechenizkiy, D.~Mocanu,
  and Z.~Wang, ``More convnets in the 2020s: Scaling up kernels beyond 51x51
  using sparsity,'' \emph{arXiv preprint arXiv:2207.03620}, 2022.

\bibitem{simonyan2014very}
K.~Simonyan and A.~Zisserman, ``Very deep convolutional networks for
  large-scale image recognition,'' \emph{arXiv preprint arXiv:1409.1556}, 2014.

\bibitem{he2016deep}
K.~He, X.~Zhang, S.~Ren, and J.~Sun, ``Deep residual learning for image
  recognition,'' in \emph{Proceedings of the IEEE conference on computer vision
  and pattern recognition}, 2016, pp. 770--778.

\bibitem{xiao2017fashion}
H.~Xiao, K.~Rasul, and R.~Vollgraf, ``Fashion-mnist: a novel image dataset for
  benchmarking machine learning algorithms,'' \emph{arXiv preprint
  arXiv:1708.07747}, 2017.

\bibitem{coates2011analysis}
A.~Coates, A.~Ng, and H.~Lee, ``An analysis of single-layer networks in
  unsupervised feature learning,'' in \emph{Proceedings of the fourteenth
  international conference on artificial intelligence and statistics}.\hskip
  1em plus 0.5em minus 0.4em\relax JMLR Workshop and Conference Proceedings,
  2011, pp. 215--223.

\bibitem{chollet2017xception}
F.~Chollet, ``Xception: Deep learning with depthwise separable convolutions,''
  in \emph{Proceedings of the IEEE conference on computer vision and pattern
  recognition}, 2017, pp. 1251--1258.

\bibitem{szegedy2015going}
C.~Szegedy, W.~Liu, Y.~Jia, P.~Sermanet, S.~Reed, D.~Anguelov, D.~Erhan,
  V.~Vanhoucke, and A.~Rabinovich, ``Going deeper with convolutions,'' in
  \emph{Proceedings of the IEEE conference on computer vision and pattern
  recognition}, 2015, pp. 1--9.

\bibitem{szegedy2016rethinking}
C.~Szegedy, V.~Vanhoucke, S.~Ioffe, J.~Shlens, and Z.~Wojna, ``Rethinking the
  inception architecture for computer vision,'' in \emph{Proceedings of the
  IEEE conference on computer vision and pattern recognition}, 2016, pp.
  2818--2826.

\bibitem{peng2017large}
C.~Peng, X.~Zhang, G.~Yu, G.~Luo, and J.~Sun, ``Large kernel matters--improve
  semantic segmentation by global convolutional network,'' in \emph{Proceedings
  of the IEEE conference on computer vision and pattern recognition}, 2017, pp.
  4353--4361.

\bibitem{hu2019local}
H.~Hu, Z.~Zhang, Z.~Xie, and S.~Lin, ``Local relation networks for image
  recognition,'' in \emph{Proceedings of the IEEE/CVF International Conference
  on Computer Vision}, 2019, pp. 3464--3473.

\bibitem{iandola2014densenet}
F.~Iandola, M.~Moskewicz, S.~Karayev, R.~Girshick, T.~Darrell, and K.~Keutzer,
  ``Densenet: Implementing efficient convnet descriptor pyramids,'' \emph{arXiv
  preprint arXiv:1404.1869}, 2014.

\bibitem{huang2018condensenet}
G.~Huang, S.~Liu, L.~Van~der Maaten, and K.~Q. Weinberger, ``Condensenet: An
  efficient densenet using learned group convolutions,'' in \emph{Proceedings
  of the IEEE conference on computer vision and pattern recognition}, 2018, pp.
  2752--2761.

\bibitem{howard2017mobilenets}
A.~G. Howard, M.~Zhu, B.~Chen, D.~Kalenichenko, W.~Wang, T.~Weyand,
  M.~Andreetto, and H.~Adam, ``Mobilenets: Efficient convolutional neural
  networks for mobile vision applications,'' \emph{arXiv preprint
  arXiv:1704.04861}, 2017.

\bibitem{zhang2018shufflenet}
X.~Zhang, X.~Zhou, M.~Lin, and J.~Sun, ``Shufflenet: An extremely efficient
  convolutional neural network for mobile devices,'' in \emph{Proceedings of
  the IEEE conference on computer vision and pattern recognition}, 2018, pp.
  6848--6856.

\bibitem{cheng2018model}
Y.~Cheng, D.~Wang, P.~Zhou, and T.~Zhang, ``Model compression and acceleration
  for deep neural networks: The principles, progress, and challenges,''
  \emph{IEEE Signal Processing Magazine}, vol.~35, no.~1, pp. 126--136, 2018.

\bibitem{vanhoucke2011improving}
V.~Vanhoucke, A.~Senior, and M.~Z. Mao, ``Improving the speed of neural
  networks on cpus,'' 2011.

\bibitem{chen2015compressing}
W.~Chen, J.~Wilson, S.~Tyree, K.~Weinberger, and Y.~Chen, ``Compressing neural
  networks with the hashing trick,'' in \emph{International conference on
  machine learning}.\hskip 1em plus 0.5em minus 0.4em\relax PMLR, 2015, pp.
  2285--2294.

\bibitem{srinivas2015data}
S.~Srinivas and R.~V. Babu, ``Data-free parameter pruning for deep neural
  networks,'' \emph{arXiv preprint arXiv:1507.06149}, 2015.

\bibitem{han2015learning}
S.~Han, J.~Pool, J.~Tran, and W.~Dally, ``Learning both weights and connections
  for efficient neural network,'' \emph{Advances in neural information
  processing systems}, vol.~28, 2015.

\bibitem{he2017channel}
Y.~He, X.~Zhang, and J.~Sun, ``Channel pruning for accelerating very deep
  neural networks,'' in \emph{Proceedings of the IEEE international conference
  on computer vision}, 2017, pp. 1389--1397.

\bibitem{gong2014compressing}
Y.~Gong, L.~Liu, M.~Yang, and L.~Bourdev, ``Compressing deep convolutional
  networks using vector quantization,'' \emph{arXiv preprint arXiv:1412.6115},
  2014.

\bibitem{wu2016quantized}
J.~Wu, C.~Leng, Y.~Wang, Q.~Hu, and J.~Cheng, ``Quantized convolutional neural
  networks for mobile devices,'' in \emph{Proceedings of the IEEE conference on
  computer vision and pattern recognition}, 2016, pp. 4820--4828.

\bibitem{liu2021post}
Z.~Liu, Y.~Wang, K.~Han, W.~Zhang, S.~Ma, and W.~Gao, ``Post-training
  quantization for vision transformer,'' \emph{Advances in Neural Information
  Processing Systems}, vol.~34, pp. 28\,092--28\,103, 2021.

\bibitem{fang2020post}
J.~Fang, A.~Shafiee, H.~Abdel-Aziz, D.~Thorsley, G.~Georgiadis, and J.~H.
  Hassoun, ``Post-training piecewise linear quantization for deep neural
  networks,'' in \emph{Computer Vision--ECCV 2020: 16th European Conference,
  Glasgow, UK, August 23--28, 2020, Proceedings, Part II 16}.\hskip 1em plus
  0.5em minus 0.4em\relax Springer, 2020, pp. 69--86.

\bibitem{li2021brecq}
Y.~Li, R.~Gong, X.~Tan, Y.~Yang, P.~Hu, Q.~Zhang, F.~Yu, W.~Wang, and S.~Gu,
  ``Brecq: Pushing the limit of post-training quantization by block
  reconstruction,'' \emph{arXiv preprint arXiv:2102.05426}, 2021.

\bibitem{zhou1994acoustic}
G.~Zhou and D.~Z. Anderson, ``Acoustic signal recognition with a
  photorefractive time-delay neural network,'' \emph{Optics letters}, vol.~19,
  no.~9, pp. 655--657, 1994.

\bibitem{larger2012photonic}
L.~Larger, M.~C. Soriano, D.~Brunner, L.~Appeltant, J.~M. Guti{\'e}rrez,
  L.~Pesquera, C.~R. Mirasso, and I.~Fischer, ``Photonic information processing
  beyond turing: an optoelectronic implementation of reservoir computing,''
  \emph{Optics express}, vol.~20, no.~3, pp. 3241--3249, 2012.

\bibitem{duport2012all}
F.~Duport, B.~Schneider, A.~Smerieri, M.~Haelterman, and S.~Massar,
  ``All-optical reservoir computing,'' \emph{Optics express}, vol.~20, no.~20,
  pp. 22\,783--22\,795, 2012.

\bibitem{jutamulia1996overview}
S.~Jutamulia and F.~Yu, ``Overview of hybrid optical neural networks,''
  \emph{Optics \& Laser Technology}, vol.~28, no.~2, pp. 59--72, 1996.

\bibitem{paquot2012optoelectronic}
Y.~Paquot, F.~Duport, A.~Smerieri, J.~Dambre, B.~Schrauwen, M.~Haelterman, and
  S.~Massar, ``Optoelectronic reservoir computing,'' \emph{Scientific reports},
  vol.~2, no.~1, p. 287, 2012.

\bibitem{woods2012photonic}
D.~Woods and T.~J. Naughton, ``Photonic neural networks,'' \emph{Nature
  Physics}, vol.~8, no.~4, pp. 257--259, 2012.

\bibitem{hughes2018training}
T.~W. Hughes, M.~Minkov, Y.~Shi, and S.~Fan, ``Training of photonic neural
  networks through in situ backpropagation and gradient measurement,''
  \emph{Optica}, vol.~5, no.~7, pp. 864--871, 2018.

\bibitem{fang2015nanoplasmonic}
Y.~Fang and M.~Sun, ``Nanoplasmonic waveguides: towards applications in
  integrated nanophotonic circuits,'' \emph{Light: Science \& Applications},
  vol.~4, no.~6, pp. e294--e294, 2015.

\bibitem{shen2017deep}
Y.~Shen, N.~C. Harris, S.~Skirlo, M.~Prabhu, T.~Baehr-Jones, M.~Hochberg,
  X.~Sun, S.~Zhao, H.~Larochelle, D.~Englund \emph{et~al.}, ``Deep learning
  with coherent nanophotonic circuits,'' \emph{Nature photonics}, vol.~11,
  no.~7, pp. 441--446, 2017.

\bibitem{ovchinnikov1999diffraction}
Y.~B. Ovchinnikov, J.~M{\"u}ller, M.~Doery, E.~Vredenbregt, K.~Helmerson,
  S.~Rolston, and W.~Phillips, ``Diffraction of a released bose-einstein
  condensate by a pulsed standing light wave,'' \emph{Physical review letters},
  vol.~83, no.~2, p. 284, 1999.

\bibitem{lin2018all}
X.~Lin, Y.~Rivenson, N.~T. Yardimci, M.~Veli, Y.~Luo, M.~Jarrahi, and A.~Ozcan,
  ``All-optical machine learning using diffractive deep neural networks,''
  \emph{Science}, vol. 361, no. 6406, pp. 1004--1008, 2018.

\bibitem{george2018electrooptic}
J.~George, R.~Amin, A.~Mehrabian, J.~Khurgin, T.~El-Ghazawi, P.~R. Prucnal, and
  V.~J. Sorger, ``Electrooptic nonlinear activation functions for vector matrix
  multiplications in optical neural networks,'' in \emph{Signal Processing in
  Photonic Communications}.\hskip 1em plus 0.5em minus 0.4em\relax Optica
  Publishing Group, 2018, pp. SpW4G--3.

\bibitem{miscuglio2018all}
M.~Miscuglio, A.~Mehrabian, Z.~Hu, S.~I. Azzam, J.~George, A.~V. Kildishev,
  M.~Pelton, and V.~J. Sorger, ``All-optical nonlinear activation function for
  photonic neural networks,'' \emph{Optical Materials Express}, vol.~8, no.~12,
  pp. 3851--3863, 2018.

\bibitem{ding2021repvgg}
X.~Ding, X.~Zhang, N.~Ma, J.~Han, G.~Ding, and J.~Sun, ``Repvgg: Making
  vgg-style convnets great again,'' in \emph{Proceedings of the IEEE/CVF
  Conference on Computer Vision and Pattern Recognition}, 2021, pp.
  13\,733--13\,742.

\bibitem{kabir2022spinalnet}
H.~D. Kabir, M.~Abdar, A.~Khosravi, S.~M.~J. Jalali, A.~F. Atiya, S.~Nahavandi,
  and D.~Srinivasan, ``Spinalnet: Deep neural network with gradual input,''
  \emph{IEEE Transactions on Artificial Intelligence}, 2022.

\end{thebibliography}

\end{document}